%% file: vilp.tex
\documentclass[doublecolumn]{IEEEtran}

\include{before_document}

\usepackage{graphicx}
\usepackage{caption}
\usepackage{verbatim}
\usepackage{multirow}
\begin{document}
\title{VILP: Imitation Learning with Latent Video Planning}
\author{
Zhengtong Xu, Qiang Qiu, Yu She$^{*}$
\thanks{$^{*}$Address all correspondence to this author.}
\thanks{Zhengtong Xu and Yu She are with the School of Industrial Engineering, Purdue University, West Lafayette, USA  (E-mail: \{xu1703, shey\}@purdue.edu).}
\thanks{Qiang Qiu is with the Elmore Family School of Electrical and Computer Engineering, Purdue University, West Lafayette, USA  (E-mail: qqiu@purdue.edu).}
}

\maketitle

\begin{abstract}
In the era of generative AI, integrating video generation models into robotics opens new possibilities for the general-purpose robot agent. This paper introduces imitation learning with latent video planning (VILP). We propose a latent video diffusion model to generate predictive robot videos that adhere to temporal consistency to a good degree. Our method is able to generate highly time-aligned videos from multiple views, which is crucial for robot policy learning.  Our video generation model is highly time-efficient. For example, it can generate videos from two distinct perspectives, each consisting of six frames with a resolution of 96x160 pixels, at a rate of 5 Hz. In the experiments, we demonstrate that VILP outperforms the existing video generation robot policy across several metrics: training costs, inference speed, temporal consistency of generated videos, and the performance of the policy. We also compared our method with other imitation learning methods. Our findings indicate that VILP can rely less on extensive high-quality task-specific robot action data while still maintaining robust performance. {In addition, VILP possesses robust capabilities in representing multi-modal action distributions.}  Our paper provides a practical example of how to effectively integrate video generation models into robot policies, potentially offering insights for related fields and directions. {For more details, please refer to our open-source repository https://github.com/ZhengtongXu/VILP.}
\end{abstract}


\section{Introduction}

With the development of generative models, video synthesis has made remarkable progress \cite{brooks2024video}. The aim of video synthesis is to generate a video clip that aligns with a given condition, such as a language prompt, an initial image, or other types of conditioning. These videos can often simulate the real world to some extent \cite{brooks2024video}. As such, video generation models are capable of predicting future interactions between objects based on the given conditioning to some degree. 

Recently, the application of video generation in robotics has garnered increasing attention. One reason for this is that, as previously mentioned, video generation models can simulate the real world to a certain extent and make predictions about the future, which aligns with the concepts of ``model" and ``planning" in robotics. Therefore, applying video generation to robotics is a natural and promising idea. Additionally, video data are often more readily available and easier to scale up compared to robot action and state data. Therefore, research on video generation in robotics may also contribute to achieving the ``scaling law" for robotics.

\begin{figure}[t]
\centering
\begin{overpic}[trim=0 0 0 0,clip, width=0.5\textwidth]{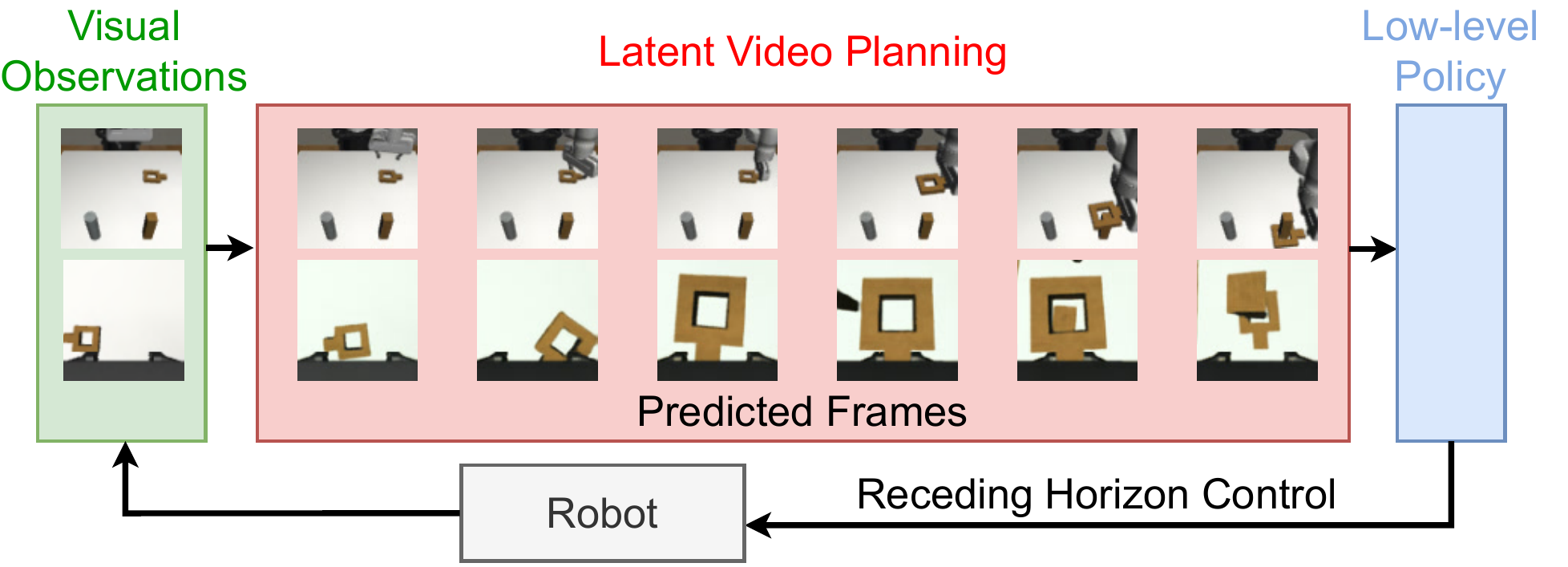}
\end{overpic}
\caption{VILP is capable of generating predictive robot videos that adhere to temporal consistency to a good degree. Within the VILP framework, these videos can be further mapped to robot actions. Through extensive experiments, we demonstrate the effectiveness and real-time performance of VILP. The video plans depicted in this figure are some frames extracted from videos produced by VILP.}
\label{fig:firstPage}
\end{figure}

One method of applying the video generation model in robotics is using it as a planner to generate videos that depict robots performing specific tasks, which can be called video planning \cite{du2023video}. Subsequently, the generated videos can be mapped to robot actions. Existing research explored this paradigm, such as learning a model from video frames to actions \cite{du2024learning,du2023video} and using optical flow to obtain actions \cite{Ko2023Learning}. However, there are still many unresolved or unexplored questions in this field. We identify three critical aspects as follows:

1. \textbf{Receding horizon planning}: due to the lengthy time required for video generation, it is often not feasible to achieve real-time performance. The works in \cite{du2024learning,du2023video,Ko2023Learning} adopt open-loop planning, or there are prolonged intervals between each replanning. In practice, planning overly long videos and generating actions based on them can lead to the accumulation of errors, ultimately causing the task to fail. Therefore, the question arises: {Can we accelerate the speed of video planning to enable real-time receding horizon planning?}

2. \textbf{Bridging the gap between video and action}: 
Given that the video generation model can be trained exclusively on video data, it is essential to optimize the use of generated videos for action generation. The model, when pretrained on task-specific videos, inherently possesses plenty of task-relevant information. This raises an intriguing question: Can we use data that vary from the tasks depicted in the videos to effectively bridge the gap between video and action?

3. {\textbf{Uniqueness of robotic data}: Due to the unique characteristics of robotic data, which includes multi-modal action distributions and multiple camera inputs, it differs from general single-view video generation tasks that do not consider multi-modality of the data \cite{ge2022long,yu2023magvit,yan2021videogpt}. Moreover, video generation is just the first step; there remains a notable gap between generating videos and translating them into actions. The design of a video generation policy that effectively adapts to the unique characteristics of robotic data is crucial. }

To address and explore the above-mentioned questions, we propose VILP, imitation learning with latent video planning, as shown in Fig.~\ref{fig:firstPage}. Our contribution can be summarized as follows.

1. {We provide a practical example that demonstrates how to effectively integrate the video generation model into policies. This integration allows VILP to represent multi-modal action distributions and operate effectively with less reliance on extensive high-quality task-specific robot action data.}

2. {VILP outperforms the existing robot video planning method UniPi \cite{du2024learning} across several metrics: training costs, inference speed, success rate, and time consistency of the generated videos. Moreover, our method is able to generate highly time-aligned videos from multiple views, which is important for robot policy.}

3. {VILP can achieve real-time receding horizon planning for our tested tasks and datasets. For instance, in one experimental task, each VILP inference generates a five-frame video of 96x160 resolution and an action sequence, completing in just 0.073 seconds per inference (approximately 14 Hz). {To the best of our knowledge, VILP is the first work capable of real-time video generation for robotic policies.}}

\section{Related Work}\label{sec:related_work}
\subsection{Imitation Learning}
Due to the simplicity in formulation and demonstrated potential in robotics, imitation learning has garnered significant attention in recent years and has seen development across various domains. For example, the works in \cite{florence2022implicit,chi2023diffusionpolicy,lee2024behavior} focus on handling the multimodal distribution in the demonstration data. A recent survey of imitation learning is presented in \cite{urain2024deep}.

Imitation learning typically requires substantial action data to achieve high-performance policies. However, videos, rich in information and more economical to collect, offer an alternative. Recent advancements in video generation models, as shown in \cite{brooks2024video}, confirm the potential of using videos to predict robotic tasks. Our paper emphasizes the integration of video generation models into policies, showing that they can achieve robust performance with less reliance on extensive, high-quality action data.

\subsection{Video Generation Model in Robotics}
Video generation models align well with temporal information, as it requires predicting temporal consistency between frames. For example, the works in \cite{du2023video,du2024learning,Ko2023Learning,yang2023learning,gu2023seer,liang2024dreamitate} propose different video generation models for robot learning. Traditionally, video generation models have required significant computational time, posing challenges for real-time robotics applications. The works mentioned above use an  ``open-loop" strategy with long replanning intervals or lack policy implementation. In contrast, through extensive experiments, we show that VILP not only produces high-quality videos but also achieves time efficiency, enabling real-time receding horizon planning.

\section{Background: Diffusion Models}

{Diffusion models \cite{sohl2015deep} are a type of generative models that learn a data distribution $p(x)$ by gradually denoising a normally distributed variable. Denote $x$ as the data (image, video, motion, etc.) that we aim to generate using the diffusion model. Starting from $x^K \sim \mathcal{N}(0, I)$, it iteratively performs $K$ denoising steps to generate a sequence of intermediate variables, $ x^{K-1},x^{K-2} \ldots, x^0$, with progressively reduced noise levels, where $x^0$ is the desired noise-free output \cite{ho2020denoising}. In this paper, we use denoising diffusion implicit models (DDIM) \cite{song2020denoising} to manage the denoising process, which can use fewer iterations to generate high-quality output.}

{The training process for DDIM is as follows. Starting from the noise-free $x^0$, train a denoising model {$\epsilon_\theta(x^k, k), k = 1, \ldots, K$,} to predict a random noise $\epsilon^k$ which was added to generate the noisy data $x^k$ at step $k$. This process can be formalized as minimizing the expected difference between the true noise $\epsilon^k$ and the noise predicted by the model $\epsilon_\theta$ across all diffusion steps. The overall objective is to reduce the mean squared error between the actual noise and the predicted noise, given by:
\begin{align}
    \mathcal{L}(\theta) = \mathbb{E}_{x^0, \epsilon^k, k}\left[ \| \epsilon^k - \epsilon_\theta(x^k, k) \|^2 \right],\label{eq:dm_loss}
\end{align}
where $x^k = \sqrt{{\alpha}^k} x^0 + \sqrt{1-{\alpha}^k} \epsilon^k$ represents the noisy version of $x^0$ at step $k$, with ${\alpha}^k$ denoting the variance schedule of the noise added over the diffusion process. This approach enables the model to learn how to reverse the diffusion process, thereby generating noise-free data $x^0$ from a sample initialized with Gaussian noise $x^K \sim \mathcal{N}(0, I)$.}

{We would like to emphasize that in this context, $x$ represents a general expression for a data point. In practice, $x$ can take various forms and represent different applications. For example, in image diffusion models \cite{ho2020denoising}, $x$ represents an image, while in diffusion policy \cite{chi2023diffusionpolicy}, $x$ denotes an action sequence.}

\section{Latent Video Planning}

\begin{figure}[t]
\centering
\begin{overpic}[trim=50 0 130 0,clip, width=0.5\textwidth]{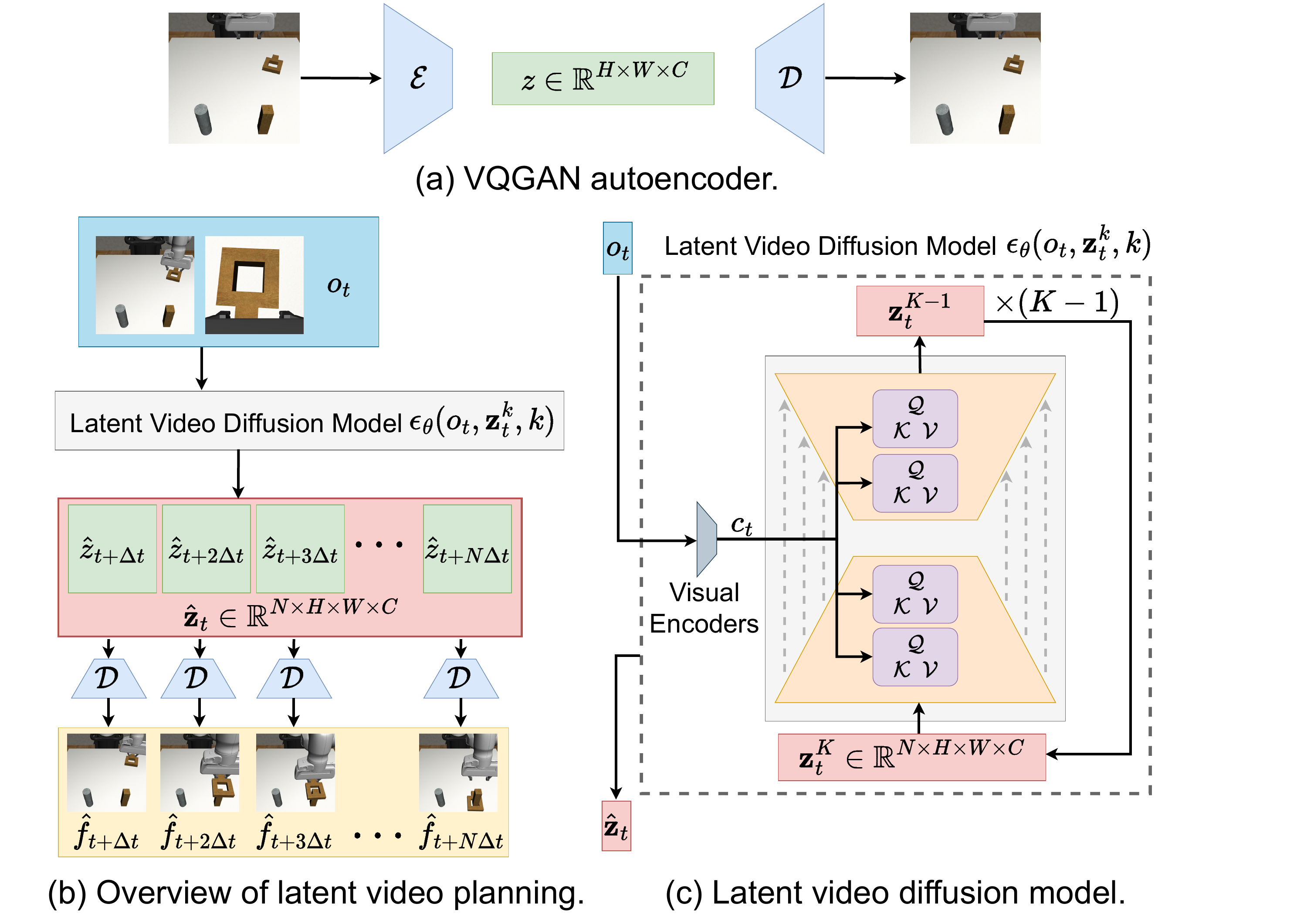}
\end{overpic}
\caption{Illustration of our proposed video planning pipeline. For the architecture of $\epsilon_\theta$, we employ a UNet model with 3D convolution layers. We adopt the cross-attention conditioning mechanism proposed in \cite{rombach2022high} and extend it to visual conditioning for 3D video data. $k$ represents the step index of the denoising process, where $k = 1, \ldots, K$. Variables with $\hat{}$ indicate the estimated/predicted variables produced by the model.
}
\label{fig:video_planning}
\end{figure}

In this section, we introduce our proposed latent video planning. The goal of video planning is given the current observation, the video planner can generate the future frames which imagine the robot execute the right actions to achieve the task. 

\subsection{Dataset Format}

In this paper, we consider video datasets with the following format $\mathcal{D}^{\text{v}}=\left\{\left(o_0^i, o_1^i, \ldots, o_{T^i_d}^i\right)\right\}_{i=1}^{E},$
where $T_d$ is the length of a video and $E$ is the amount of episodes. $o$ is a combination of different images from multiview or RGBD cameras. Denote $o = \{f^j\}_{j=1}^M$, where $f$ is one image frame and $M$ is the total count of views included. 

\subsection{Image Compression to Latent Space}\label{sec:cprs}
We use VQGAN autoencoder proposed in \cite{rombach2022high,esser2021taming} as our compression model. It is trained by a combination of perceptual loss \cite{zhang2018unreasonable} and patch-based adversarial objective \cite{yu2021vector,esser2021taming}, with the aim of maintaining local realism, reducing blurriness, and avoiding excessively high-variance latent spaces.

As shown in Fig.~\ref{fig:video_planning}(a), given an image $f \in \mathbb{R}^{\tilde{H} \times \tilde{W}\times \tilde{C}}$, the encoder $\mathcal{E}$ compresses $f$ into a latent representation $z = \mathcal{E}(f) \in \mathbb{R}^{H \times W \times C}$, and the decoder $\mathcal{D}$ reconstructs the image from $z$. $f$ here can be a RGB image ($\tilde{C}=3$) or a depth image ($\tilde{C}=1$).

{Once the VQGAN autoencoder is trained with the objective of image compression, it is fixed and not adjusted during the subsequent training of the diffusion model. }

\subsection{Latent Video Diffusion Model}

While latent diffusion model \cite{rombach2022high} is for image synthesis, we use the idea of diffusion in latent space for video synthesis. First, sample a sequence of frames for the dataset $\mathcal{D}^{\text{v}}$, denoted as 
$$\mathbf{f}_t = [f_{t+\Delta t},f_{t+2\Delta t},\ldots,f_{t+N\Delta t}]\in \ \mathbb{R}^{N \times \tilde{H} \times \tilde{W} \times \tilde{C}},$$ 
where $N$ is the sampling length and $\Delta t$ is the sampling interval. By the compression model described in Section~\ref{sec:cprs}, we can compress $\mathbf{f}_t$ to a latent variable
$$\mathbf{z}_t = [z_{t+\Delta t},z_{t+2\Delta t},\ldots,z_{t+N\Delta t}]\in \ \mathbb{R}^{N \times H \times W \times C},$$
where $z = \mathcal{E}(f) \in \mathbb{R}^{H \times W \times C}$. To generate $\mathbf{f}_t$ by diffusion models, we can generate $\mathbf{z}_t$ first, and use the decoder $\mathcal{D}$ to reconstruct the video. Therefore, replace $x$ with $\mathbf{z}_t$ in equation \eqref{eq:dm_loss}, and the objective of our proposed video diffusion model is given by
\begin{align}
    \mathcal{L}(\theta) = \mathbb{E}_{\mathbf{z}^0_t, \epsilon^k, k}\left[ \| \epsilon^k - \epsilon_\theta(\mathbf{z}^k_t, k) \|^2 \right].
    \label{eq:video_loss}
\end{align}

For the architecture of $\epsilon_\theta(x^k, k)$, we employ a UNet model \cite{dhariwal2021diffusion,ronneberger2015u} with 3D convolution layers. This configuration enables the model to effectively capture and integrate both temporal and spatial information. 

\subsection{Observation Conditioning and Multiview Generation}
To make the latent video diffusion model serve as a video planner, we design the UNet model conditioned on observation $o_t$. In this way, the diffusion model approximates the conditional distribution $p(\mathbf{z}_t|o_t)$. Therefore, extending equation \eqref{eq:video_loss}, the objective of our proposed latent video planner is given by 
\begin{align*}
    \mathcal{L}(\theta) = \mathbb{E}_{(o_t,\mathbf{z}^0_t),\epsilon^k,k}\left[ \| \epsilon^k - \epsilon_\theta(o_t,\mathbf{z}^k_t, k) \|^2 \right].
\end{align*}

Specifically, the VILP conditioning mechanism comprises three parts: 

1. The first part involves mapping the images in the observation $o$ to low-dimensional vectors using visual encoders. We adopt the same encoder used in \cite{chi2023diffusionpolicy}, which is a modified ResNet-18 (without pretraining). In our implementation, different camera perspectives utilize separate encoders. For depth image inputs, we process them by repeating the depth image three times to create a three-channel image and then input to the encoder. Ultimately, we concatenate all the output from the encoders into a low-dimensional vector {$c_t \in \mathbb{R}^{D_c}$}. 

The visual encoders discussed are designed differently from those used for image compression. Encoders for image compression aim to produce low-variance latent spaces and ensure that reconstructed images maintain local realism and minimize blurriness. In contrast, encoders used for conditioning focus on maintaining stable training across the diffusion pipeline and capturing detailed scene information.

2. The second part addresses how to embed $c_t$ for conditioned video generation. We adopt the conditioning mechanism proposed in \cite{rombach2022high} and extend it to visual conditioning for 3D video data. Specifically, this involves mapping $c_t$ to the intermediate layers of the 3D UNet via a cross-attention layer implementing $\text{Attention}(\mathcal{Q}, \mathcal{K}, \mathcal{V})=\operatorname{softmax}\left(\frac{\mathcal{Q} \mathcal{K}}{\sqrt{d}}\right) \cdot \mathcal{V}$, where 
$$\mathcal{Q} =\mathcal{W}_\mathcal{Q} ^{h} \cdot \varphi^h\left(\mathbf{z}_t^k\right), \mathcal{K}=\mathcal{W}_\mathcal{K}^{h} \cdot c_t, \mathcal{V}=\mathcal{W}_\mathcal{V}^{h} \cdot c_t.$$
Here, $h$ represents the index of a head in the multi-head attention mechanism and  $\varphi^h\left(\mathbf{z}_t^k\right)$ denotes an intermediate representation of the UNet $\epsilon_\theta$. Given that \( \mathbf{z}_t^k \) is a vector with temporal and spatial features, \( \varphi^h \) simultaneously flattens both the temporal and spatial dimensions.
 $\mathcal{W}_\mathcal{Q} ^{h}, \mathcal{W}_\mathcal{K}^{h}$, and $\mathcal{W}_\mathcal{V}^{h}$ are learnable transformations \cite{jaegle2021perceiver,vaswani2023attention}.

3. For certain applications, observations from multiview are essential as they enable more effective conditioning, enhancing video and action generation that are challenging for single-perspective inputs. Therefore, for video planning, we train a diffusion model from scratch for each view's image to facilitate the generation of multiview videos, resulting in $\{\epsilon^j_\theta(o_t,\mathbf{z}^k_{t}
,k)\}_{j=1}^M$, where $M$ is the count of views. In addition, $c_t$ requires the integration of embeddings from multiview observations, as depicted in Fig.~\ref{fig:video_planning}(b). This conditioning method make generated videos of different views align well temporally with each other. 

We would like to emphasize that for imitation learning datasets, having multiview image observations is very common \cite{chi2023diffusionpolicy,xu2024unit}. Although training diffusion models for each view increases training costs, it aligns well with imitation learning datasets. Our pipeline is an effective way to generates time-aligned robot videos across multiple views.

\begin{figure}[t]
\centering
\begin{overpic}[trim=40 0 0 0,clip, width=0.45\textwidth]{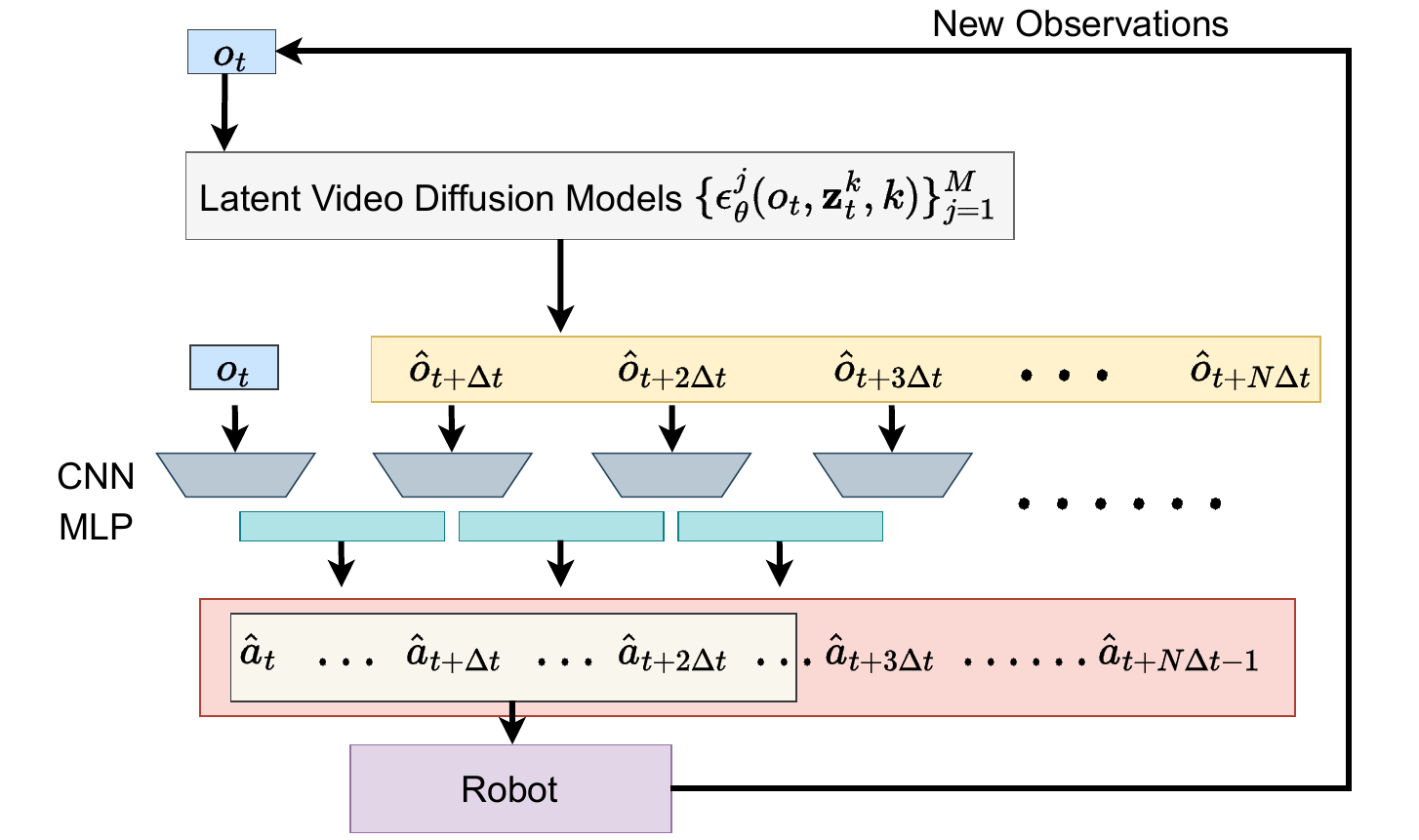}
\end{overpic}
\caption{Overview of the low-level policy that maps the predicted video to predicted action sequence.}
\label{fig:low_level_policy}
\end{figure}

\section{Imitation Learning with Video Planning}
In this section, we will detail how to use the planned future frames to generate actions. We use a goal-conditioned policy \cite{du2023video} as our low-level policy.

The overview of the low-level policy is shown in Fig.~\ref{fig:low_level_policy}. We use two adjacent generated frames to generate an action sequence. Denote the action mapping module as 
$$\hat{\mathbf{a}}_{t+n\Delta t} = \pi (\hat{o}_{t+n\Delta t},\hat{{o}}_{t+(n+1)\Delta t}), n = 0,1,\ldots,N-1$$
where $\hat{\mathbf{a}}_{t+n\Delta t} = [a_{t+n\Delta t},a_{t+n\Delta t+1},\ldots,a_{t+(n+1)\Delta t -1}]$. The module consists of two CNN encoders and one MLP head. As shown in Fig.~\ref{fig:low_level_policy}, we use the same module to mapping all predicted future frames to predicted action sequences, and the action sequences are continuous from time step $t$ to $t+N\Delta t -1$.

In the end, we use a receding horizon control strategy, as shown in Fig.~\ref{fig:low_level_policy}. Instead of executing all generated actions, we execute only the first \(N_e\) steps. This approach enhances policy responsiveness and reduces cumulative errors in video and action prediction that increase with time steps.

\begin{table*}[t]
\centering
{
\begin{tabular}{cl|ccc|ccc}
\hline
 & & \ {VILP-4} & \ {VILP-8} & \ {VILP-16} &\ {UniPi-4} & \ {UniPi-16} & \ {UniPi-64} \\
\hline
\multirow{2}{*}{{Move-the-Stack}} & \ {FID} & 40.80 & 41.42 & 41.33 & 94.01 & 55.85 & \textbf{39.04}\\
                                   & \ {FVD} &479.59& 447.75 & 453.16 & 490.81& 481.34& \textbf{429.36}\\
                                  & \ {Time} & \textbf{0.058} &0.11 & 0.21 & 0.14 & 0.61 & 2.5 \\
\hline
\multirow{2}{*}{{Push-T}}         & \ {FID} & 17.85  &\textbf{14.65} & 16.49  & 158.50 & 25.43 & 16.72 \\
                                  & \ {FVD} & 454.11 & 460.98 & \textbf{447.56} & 1174.83 & 744.06 &675.85\\
                                  & \ {Time} & \textbf{0.058} & 0.11 & 0.21 &0.14 & 0.61 &2.5 \\
\hline
\multirow{2}{*}{{Towers-of-Hanoi}} & \ {FID} &20.33 & \textbf{16.20} & 16.25 & 121.68 & 52.32 & 23.63 \\
                                   & \ {FVD} &249.03&225.80 & \textbf{223.77} &570.41& 465.67 & 390.79 \\
                                   & \ {Time} & \textbf{0.187} & 0.350 & 0.677 & 0.47 & 2.0 & 8.3 \\
\hline
\end{tabular}
}
\caption{Comparative results of video planning between VILP and UniPi. {We report FIDs and FVDs \cite{yan2021videogpt} on these three tasks.  We employed a 9:1 episode split, using 90\% of the data to train VILP and UniPi, and reserving the remaining 10\% as unseen data. This unseen portion was used to evaluate the FID and FVD scores of videos generated by VILP and UniPi, conditioned on the first frame of these unseen episodes. } Time (measured in seconds, averaged over 50 inferences) refers to the duration required for a single inference with a batch size of 1, corresponding to the prediction of one video clip. Consequently, 1/Time represents the frequency at which video planning can be executed. {The GPU employed for profiling the running time is a single NVIDIA RTX A6000.}}
\label{tab:video_planning}
\end{table*}

\begin{figure}[t]
\centering
\begin{overpic}[trim=0 0 0 0,clip, width=0.5\textwidth]{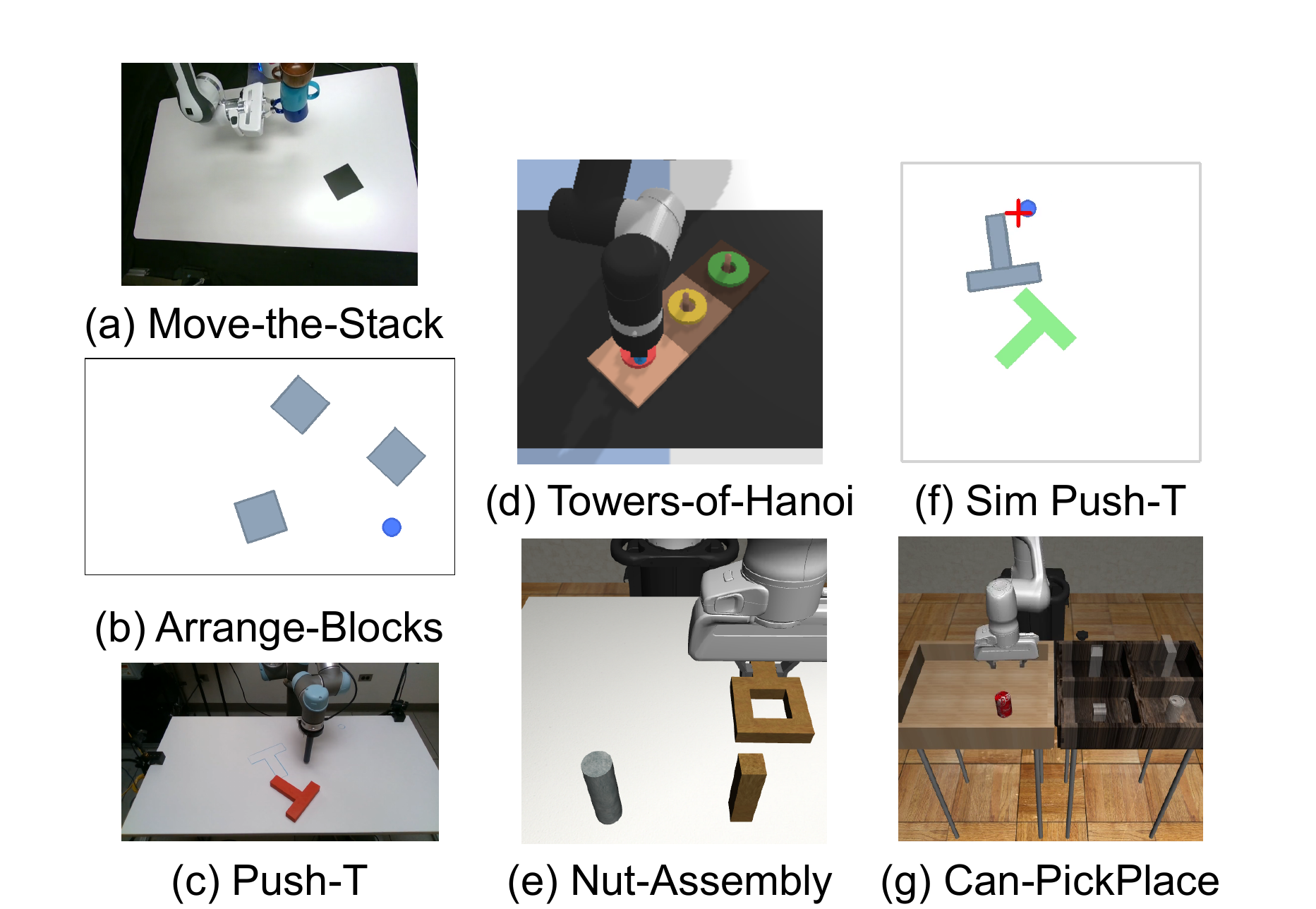}
\end{overpic}
\caption{Tasks for experiments. Move-the-Stack \cite{xu2024leto}, Push-T \cite{chi2023diffusionpolicy}, Towers-of-Hanoi \cite{zeng2021transporter} are for video planning experiments (no policy involved). {Nut-Assembly \cite{mandlekar2022matters}, Can-PickPlace \cite{mandlekar2022matters}, Sim Push-T \cite{chi2022iterative}, and Arrange-Blocks \cite{chi2022iterative} are for the evaluation of the whole process which consists of video planning and low-level policy.} Move-the-Stack and Push-T are in real environment while the rest tasks are in simulation.}
\label{fig:vp_exp}
\end{figure}

\section{Experiments}

We evaluated VILP from two perspectives: video planning only and policy rollout. Policy rollout means not only involving video planning but also translating the video planning into actions to drive the robot in task execution. In the video planning experiments (Section~\ref{sec:vp_exp}), we evaluated the effectiveness of VILP in generating future videos that ``imagine" the robot successfully executing tasks, in comparison to the video planning module of UniPi \cite{du2024learning}. In terms of policy rollout (Section~\ref{sec:rollout_exp}), we compared VILP not only with UniPi but also with diffusion policy \cite{chi2023diffusionpolicy}. Additionally, we assessed the training costs and inference speeds of VILP relative to UniPi.

Our UniPi implementation off of the UniPi implementation in \cite{Ko2023Learning}. To ensure experimental fairness, we implement the same low-level policy of VILP on UniPi to map videos to actions. Moreover, we set UniPi and VLIP employ identical base channel counts and channel multipliers within their U-Net architectures. Notably, the numbers following the methods indicate the denoising steps used. Both UniPi and VILP utilize DDIM. For instance, ``VILP-4" denotes that VILP with 4 denoising steps using DDIM. 

\subsection{Video Planning}\label{sec:vp_exp}

\begin{table}[t]
\centering
\begin{tabular}{clcc|c}
\hline
 & & \ {Frames} & \ {Size} &\ {Mem.} \\
\hline
\multirow{3}{*}{{Move-the-Stack}} & \ {VILP} & 8 & 96,128 & 8.9 GB  \\
                                 & \ {UniPi} & 5& 96,128 & 35.7 GB   \\
                                 &\ {UniPi} & 6& 96,128 & 44.1 GB   \\
\hline
\multirow{3}{*}{{Push-T}} & \ {VILP} & 6 & 96,128&  7.4 GB \\
                                 &\ {UniPi} & 5 & 96,128 & 35.7 GB  \\
                                 &\ {UniPi} & 6 & 96,128 & 44.1 GB  \\
\hline
\multirow{2}{*}{{Towers-of-Hanoi}} & \ {VILP} & 6 & 160,96 & 8.7 GB \\
                                 & \ {UniPi} & 6 & 160,96 & 68.2 GB  \\
\hline
\multirow{2}{*}{{Nut-Assembly}} & \ {VILP} & 6 & 80,80 & 6.2 GB   \\
                                 & \ {UniPi} & 6 & 80,80 & 20.5 GB  \\
\hline
\multirow{2}{*}{{Arrange-Blocks}} & \ {VILP} & 5 & 96,160 & 10.0 GB  \\
                                 & \ {UniPi} & 5 & 96,160 & 82.5 GB  \\
\hline
\end{tabular}
\caption{GPU memory occupation when training with batch size 16. }
\label{tab:video_mem}
\end{table}

We implemented VILP and UniPi across three video datasets: Move-the-Stack \cite{xu2024leto}, Push-T \cite{chi2023diffusionpolicy}, and Towers-of-Hanoi \cite{zeng2021transporter}, as shown in Fig.~\ref{fig:vp_exp}. {We report FIDs and FVDs \cite{yan2021videogpt} on these three tasks.  We employed a 9:1 episode split, using 90\% of the data to train VILP and UniPi, and reserving the remaining 10\% as unseen data. This unseen portion was used to evaluate the FID and FVD scores of videos generated by VILP and UniPi.} The first inference utilizes initial images sampled from episodes in the unseen test set, while the subsequent inferences use the last generated frames from the previous inference. This approach allows the model to generate long videos and ensures that all test cases are instances not previously seen in the training set.

The goal of the Move-the-Stack task is for the robot to relocate stacked cups to a black board. Both the position of the black board and the robot's initial position are randomized. In the Push-T task, the objective is for the robot to push a randomly placed T-block to a predetermined target position and orientation. For the Towers-of-Hanoi task, the robot is required to solve the Tower of Hanoi puzzle with three disks. {The objective of the task is moving all disks from one peg to another using a third peg, following these rules: only one disk moves at a time, no larger disk on a smaller one.} In our setup, the position of the base remains fixed, whereas the orientation of the base varies randomly between $(\frac{\pi}{4}, \frac{3\pi}{4})$.

{As shown in Table~\ref{tab:video_planning}, for the Push-T and Towers-of-Hanoi tasks, VILP achieved better FID and FVD scores than UniPi. For the Move-the-Stack task, UniPi obtained the optimal FID and FVD scores when inference steps were set to 64. However, at this setting, UniPi's inference time was substantially higher than VILP’s. VILP, in contrast, achieved comparable FID and FVD scores with a much faster inference speed. Moreover, our results show that VILP can generate high-quality videos even with fewer inference steps (e.g., 4 steps), while UniPi’s video quality decreases significantly under the same conditions. This underscores VILP's exceptional time efficiency in robot video planning.}

\begin{figure}[t]
\centering
\begin{overpic}[trim=0 0 0 0,clip, width=0.49\textwidth]{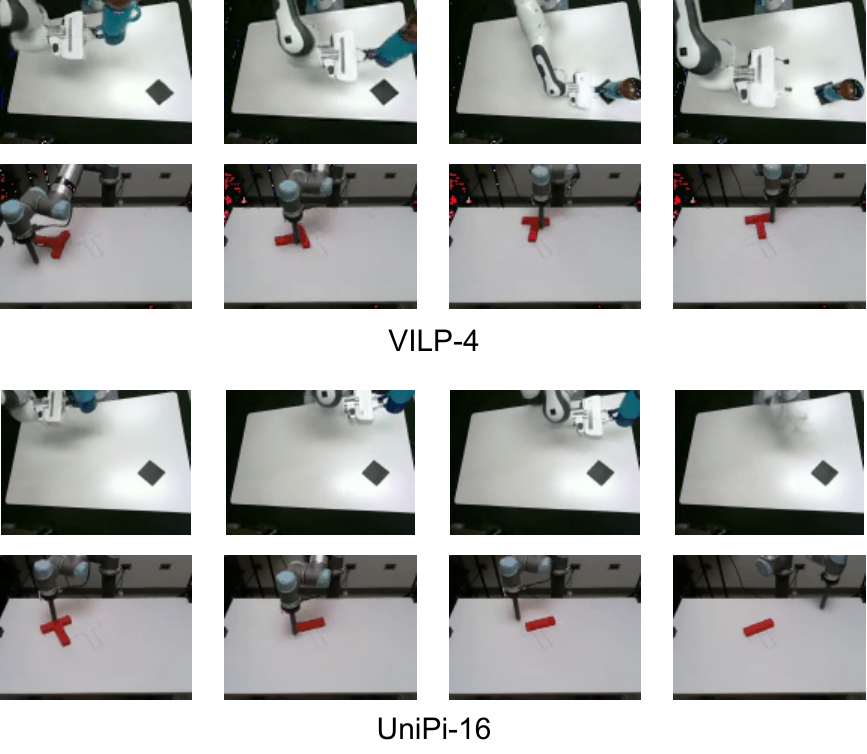}
\end{overpic}
\caption{Example frames extracted from the generated videos. For more synthesised video visualizations, please see our supplementary video. }
\label{fig:example}
\end{figure}

As shown in Fig.~\ref{fig:example}, and the supplementary video, VILP outperforms UniPi in generating video plans in terms of stronger temporal consistency and a more stable ability to capture the dynamics of robot and object movements. We attribute this enhancement to the following factors: 1. Employing diffusion model in the latent space reduces computational complexity, enabling better data distribution capture. 2. VILP's conditioning mechanism globally integrates the condition images into U-Net, whereas UniPi uses conditional concatenation. Global conditioning in robot video planning provides better control over the generation process. As shown in Table~\ref{tab:video_mem}, VILP also surpasses UniPi in terms of training costs.

\begin{figure}[t]
\centering
\begin{overpic}[trim=0 0 0 0,clip, width=0.49\textwidth]{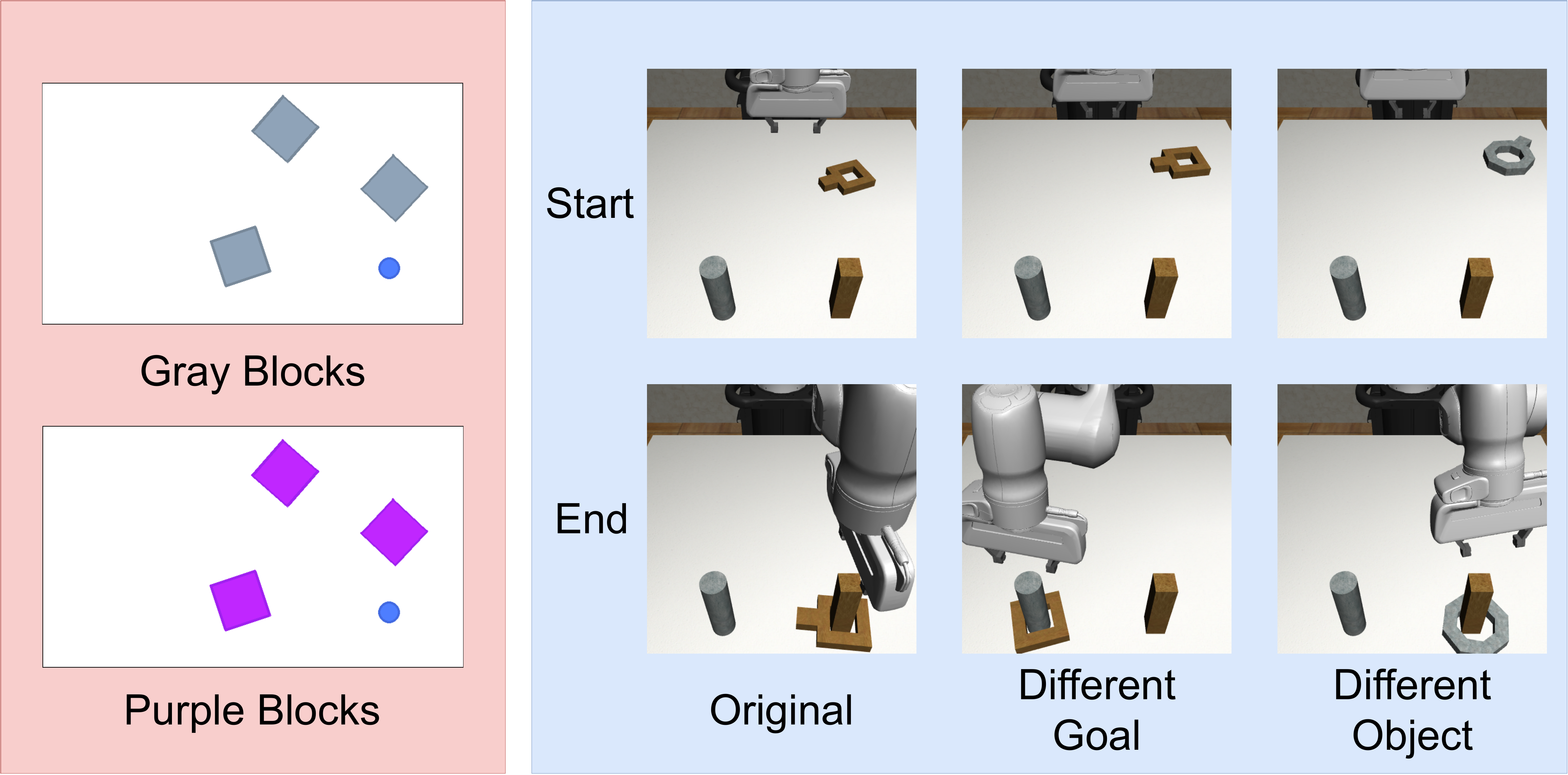}
\end{overpic}
\caption{Illustration of ``Hybrid" datasets.}
\label{fig:hybrid_square}
\end{figure}

\begin{table*}[t]
\centering
\begin{tabular}{lccccccc}
\hline
 &  \ {Epochs Rollout}&\ {Diffusion Policy} & \ {VILP-4}   & \ {UniPi-16} \\
\hline
\ {Nut-Assembly-Small} &300& $\textbf{28.0} \pm 4.0/\textbf{24.3} \pm 1.0$ & $26.7 \pm 1.2/23.3 \pm 3.4$ & $20.0 \pm 2.0/17.6 \pm 1.2$  \\
\hline
\ {Nut-Assembly-Hybrid}&100& $48.0 \pm 7.2/43.1 \pm 6.3$ & $\textbf{56.7} \pm 4.6/\textbf{53.2} \pm 6.2$ & $35.3 \pm 4.2/27.2 \pm 3.0$  \\
\hline
 & \ {Epochs Rollout}&\ {Diffusion Policy} & \ {VILP-8}   & \ {UniPi-16} \\
 \hline
\ {Arrange-Blocks-Small} &300& $8.9\pm2.0/5.9\pm0.8$ & $\textbf{46.0}\pm7.2/\textbf{40.4}\pm5.8$ & $14.7\pm1.2/8.9\pm0.2$  \\
\hline
\ {Arrange-Blocks-Hybrid} &100& $22.7 \pm 3.1/17.1 \pm 1.2$ & $\textbf{84.0} \pm 0.0/\textbf{77.6} \pm 2.8$ & $18.7 \pm 2.3/16.2 \pm 1.7$ \\
\hline
\end{tabular}
\caption{Success rate of Nut-Assembly and Arrange-Blocks. We report {average of 3 seeds $\pm$ std} results on 3$\times$50 environments. The results here are {max/mean} success rate. Max success rate is the highest success rate among the training process while mean success rate is the average of five consecutive policy rollouts (we report the highest average among the training). Epochs Rollout represents training epochs between each two rollouts.}
\label{tab:model_success_rates}
\end{table*}

\begin{table}[t]
\centering
\begin{tabular}{lcccccc}
\hline
 &  \multicolumn{3}{c}{{Nut-Assembly-Hybrid}}& \\
\hline
 & \ {VILP-4} & \ {VILP-8} & \ {VILP-16} &\ {UniPi-16} \\
\hline
\ {Score}  & $54.0/\textbf{50.0}$ & $52.0/48.0$ & $\textbf{60.0}/48.0$ & $40.0/30.0$  \\
\ {Time} & $\textbf{0.148}$& $0.257$& $0.465$  & $0.709$  \\
\hline
 &  \multicolumn{3}{c}{{Arrange-Blocks-Hybrid}}& \\
\hline
 & \ {VILP-4} & \ {VILP-8} & \ {VILP-16} &\ {UniPi-16} \\
\hline
\ {Score} & $76.0/64.8$ & $\textbf{84.0/80.4}$ & $\textbf{84.0}/80.0$ & $18.0/13.2$  \\
\ {Time}  & $\textbf{0.073}$ & $0.125$ & $0.231$ & $1.43$  \\
\hline
\end{tabular}
\caption{Ablation study of denoising steps for VILP and UniPi. The scores are {max/mean} success rates calculated in the same manner as in Table~\ref{tab:model_success_rates}. Time (measured in seconds, averaged over 50 inferences) refers to the duration required for a complete single inference, including video and action generation. Therefore, 1/Time represents the frequency at which the whole policy can be executed.}
\label{tab:denoising_benchmark}
\end{table}

\begin{table*}[t]
\centering
{
\begin{tabular}{|c|c|c|c|c|c|c|c|}
\hline
         &{UniPi} & DiffusionPolicy-C & DiffusionPolicy-T & LSTM-GMM & IBC & VILP w/o low. dim. & VILP w/ low. dim. \\ \hline
S. Push-T & {82.0}& 84                & 66                & 54       & 64  & 82.6               & \textbf{88.0}              \\
Can       & {37.4}& 97                & \textbf{98}                & 88       & 1   & 95.7               & 92.2              \\ \hline
\end{tabular}}
\caption{{{The results of the comparison between Unipi, VILP, and diffusion policy (``C" and ``T" denote backbones based on CNN and transformer) \cite{chi2023diffusionpolicy}, LSTM-GMM \cite{mandlekar2022matters}, and IBC \cite{florence2022implicit}, on the Sim Push-T and Can-PickPlace tasks. Both UniPi and VILP use 16 inference steps.} All policies in this study are visual policies rather than exclusively low-dimensional ones. In addition to image observations, VILP labeled with ``w/ low. dim." also use low-dimensional observations. We present success rates with average of last 10 checkpoints, with each averaged across 3 training seeds and 50 different environment initial conditions (150 in total). We used the same seeds and experimental setup as those in \cite{chi2023diffusionpolicy}; therefore, the baseline results are directly adapted from \cite{chi2023diffusionpolicy}.}}
\label{tab:vilp_bc_compare}
\end{table*}

\begin{table}[t]
\centering
{
\begin{tabular}{|c|c|c|c|c|}
\hline
           & cond. cat. & w/o fusion & w/o low dim. & w/ low dim. \\ \hline
S. Push-T & 72.5                 & -                          & {82.6}             & \textbf{88.0}              \\
Can        & 61.7                 & 60.4                       & \textbf{95.7}             & {92.2}            \\ \hline
\end{tabular}}
\caption{{The results of model module ablation studies. Performance are reported in the same format as in Table~\ref{tab:vilp_bc_compare}. All methods benchmarked here involve diffusion and denoising in the latent space, and are visual policies. cond. cat. represents conditional concatenation. This conditioning mechanism concatenates the conditional image with noise, which is then jointly input into the U-Net (a technique also employed by UniPi \cite{du2024learning} and AVDC \cite{Ko2023Learning}). The term ``w/o fusion" refers to using only the observation from the generated perspective as the condition, rather than performing fusion across multiple perspectives. Since the observations in Sim Push-T are from a single viewpoint, this method does not apply.}}
\label{tab:ablation_vilp}
\end{table}

\begin{figure}[t]
\centering
\begin{overpic}[trim=0 0 0 0,clip, width=0.5\textwidth]{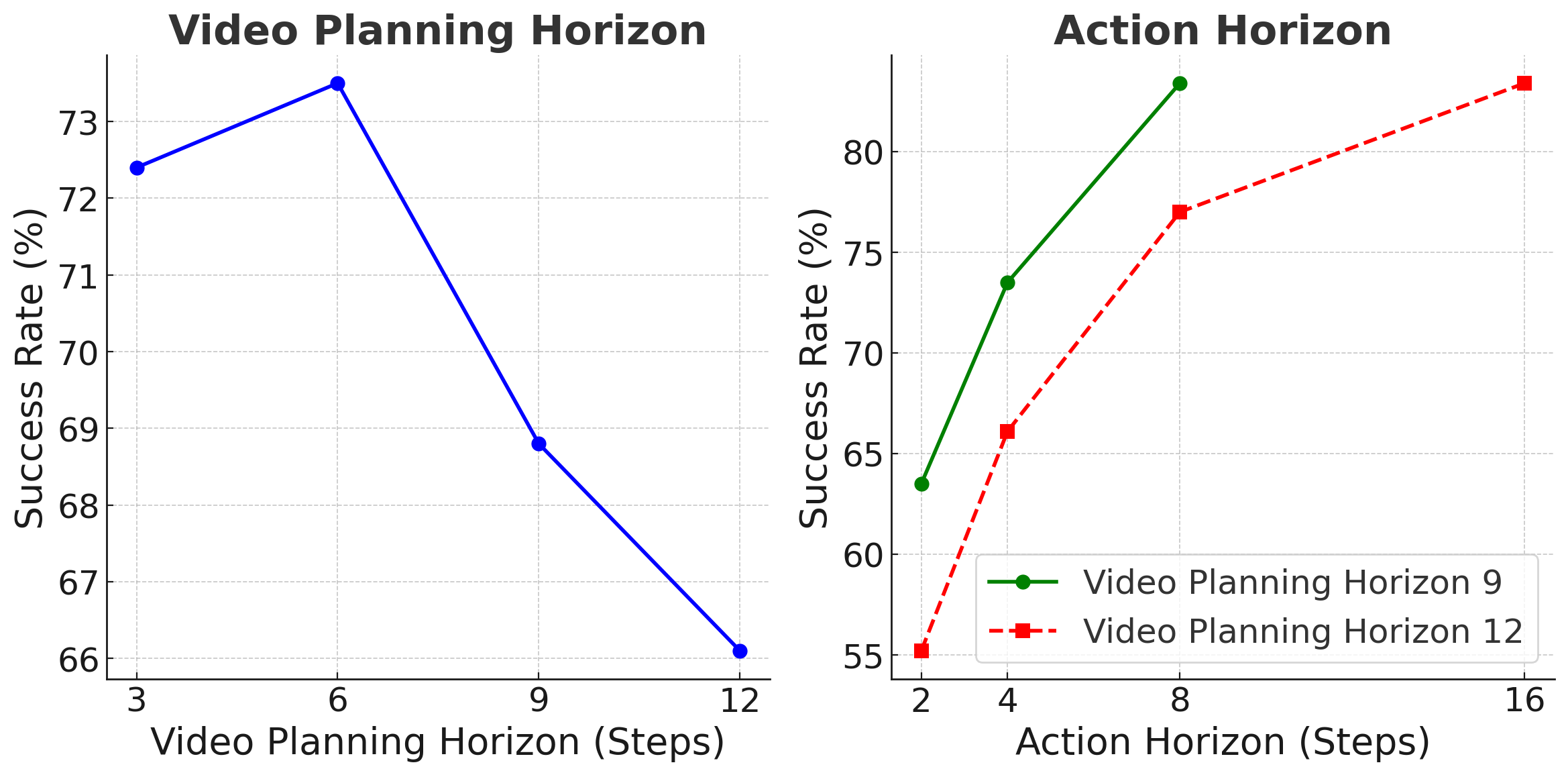}
\end{overpic}
\caption{{Horizon Ablation Study: The results are based on Sim Push-T task. The video planning horizon refers to the length of the predicted video, while the action horizon is the length of the executed action. The video planning horizon ablation is evaluated under an action horizon of 4. A larger video planning horizon allows the low-level policy to convert the video into a action sequence with longer length, thereby extending the maximum value of the action horizon.}
}
\label{fig:horizon_ablation}
\end{figure}

\subsection{Imitation Learning with Video Planning}\label{sec:rollout_exp}
\subsubsection{Simulation Tasks}

In this section, we present the results of implementing VILP as policies in four simulation tasks: Nut-Assembly \cite{mandlekar2022matters}, Can-PickPlace\cite{mandlekar2022matters}, Sim Push-T \cite{chi2022iterative}, and Arrange-Blocks \cite{chi2022iterative}. All policies in this section are visual policies.

The VILP and UniPi video planning models are trained using datasets of 610 and 200 video episodes for the Arrange-Blocks and Nut-Assembly tasks, respectively. For low-level policy training, we designed two action-labeled data configurations for each task, denoted as ``Small" and ``Hybrid". For Nut-Assembly, the ``Small" configuration comprises 10 action-labeled demonstrations, whereas the ``Hybrid" configuration expands this dataset to include an additional 75 action-labeled demonstrations with different goals and 75 action-labeled demonstrations with different objects, supplementing the original ``Small" set, as illustrated in Fig.~\ref{fig:hybrid_square}. In a similar vein, Arrange-Blocks-Small consists of 16 action-labeled demonstrations using gray blocks, and Arrange-Blocks-Hybrid enriches this with 60 more action-labeled demonstrations featuring purple blocks, also shown in Fig.~\ref{fig:hybrid_square}. Diffusion policy is trained on action-labeled data only.

From our experimental results, we can draw the following conclusions:

1. Echoing insights from video planning comparisons, VILP exhibits several advantages over UniPi, including lower training costs, faster inference times, and superior video generation quality. Consequently, VILP outperforms UniPi in practical policy applications. {Predicting in latent space is the main reason for fast inference. Due to the role of image compression, the size of the latent space is much smaller than the original pixel space. For example, an image of 96x160 pixels can be compressed into a 12x20 space.}

2. VILP and diffusion policy focus on different aspects. The comparison here merely aims to illustrate that when we have access to a large amount of video data for a task but scarce or disorganized action-labeled data, VILP can perform exceptionally well, as shown in Tables~\ref{tab:model_success_rates} and \ref{tab:denoising_benchmark}. This is because video generation models trained on video data provide a lot of information about the task. 
We demonstrate that an efficient video-to-action model can be developed using a small action dataset combined with off-target demonstrations, suggesting that training for video and action models can be  separated. Traditional imitation learning relies heavily on action-labeled data, but with adequate high-quality action data, methods like diffusion policy can be effective.

3. {Sometimes, the videos generated by VILP contain minor artifacts. Fortunately, the experimental results indicate that these do not adversely affect VILP's performance as a policy. This suggests that despite some noise and artifacts, the generated videos still effectively simulate interactions between robots and objects.}

{We also compared VILP with other imitation learning methods on the Sim Push-T and Can-PickPlace tasks. Here, we trained VILP entirely in the imitation learning data format, where all videos were paired with high-quality action labels. The results are shown in Table~\ref{tab:vilp_bc_compare}. All outcomes presented in Table~\ref{tab:vilp_bc_compare} are from visual imitation learning. It can be seen that VILP demonstrated results on the Can-PickPlace task that were comparable to those of the diffusion policy and superior to other methods. Moreover, VILP showed the best results on the Sim Push-T task. Given the strong action multi-modal distribution present in the Sim Push-T task, this indicates that VILP also possesses robust capabilities in representing multi-modal distributions. This strength stems from the generative nature of the diffusion model on which the video generation is based.}

{
Moreover, we conducted an ablation study of VILP on the Sim Push-T and Can-PickPlace tasks, with the results presented in Table~\ref{tab:ablation_vilp}. These results clearly demonstrate the effectiveness of VILP's conditioning mechanism and its handling of multi-view observations and their generation.}

{Finally, we conducted an ablation study on the Sim Push-T task concerning the video planning horizon and action horizon, with the results depicted in Fig.~\ref{fig:horizon_ablation}. This study provides valuable guidance on selecting the horizon parameters for VILP.} 

{The results indicate that both excessively small and large video horizons can negatively impact VILP performance. Additionally, increasing the action horizon tends to enhance VILP performance. Notably, a larger video planning horizon allows the low-level policy to convert the video into a longer action sequence, thereby extending the maximum value of the action horizon. Thus, sometimes, enlarging the video planning horizon can result in better performance due to the allowance of a larger action horizon. However, due to the high training and inference costs associated with video planning, a very large video planning horizon is clearly impractical. Therefore, in practical parameter selection, it is essential to trade off these factors. Based on our experimental results, optimal parameters include a video planning horizon of 6 combined with an action horizon of 8, or a video planning horizon of 12 paired with an action horizon of 16.}

\begin{figure*}[t]
\centering
\begin{overpic}[trim=0 0 0 0,clip, width=0.98\textwidth]{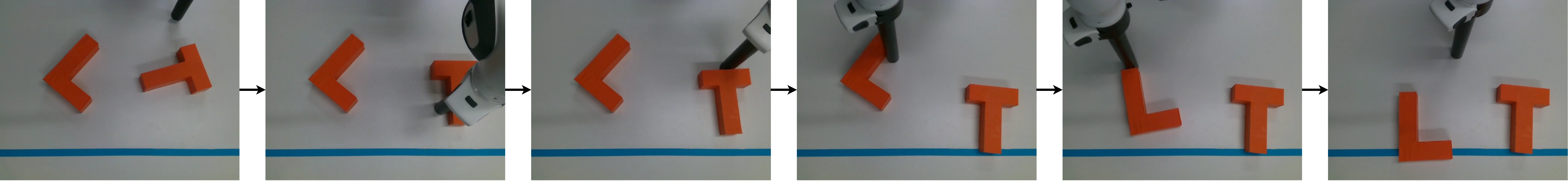}
\end{overpic}
\caption{{Snapshots of the VILP rollout in the Real-Arrange-Blocks task.}}
\label{fig:real_exp}
\end{figure*}

\begin{table}[t]
\centering
{
\begin{tabular}{|c|c|c|c|}
\hline
                    & VILP-16 & UniPi-4 & UniPi-16 \\ \hline
Two Blcoks Arranged & \textbf{7/15}    & 0/15    & 0/15     \\ \cline{1-1}
One Block Arranged  & \textbf{6/15}   & 0/15    & 0/15     \\ \cline{1-1}
Inference Time      & \textbf{0.238}   & 0.358   & 1.422    \\ \hline
\end{tabular}}
\caption{{Experimental results of the Real-Arrange-Blocks task.``Two Blocks Arranged" indicates the success rate of completing the entire task, while ``One Block Arranged" represents the rate at which only one block is arranged, marking a failure case. However, this metric also provides insight into the performance of the policies. Inference time (measured in seconds, averaged over 50 inferences) refers to the duration needed for a complete single inference, including both video and action generation. It is worth noting that the inference times for this real-world task are similar with those observed for simulation tasks, as shown in Table~\ref{tab:denoising_benchmark}.}}
\label{tab:real_sr}
\end{table}

\subsubsection{Real Environment Task}
{We compare VILP and UniPi on a real-world task, Real-Arrange-Blocks. The task goal is to arrange two blocks, an L block and a T block, on a blue line in an "LT" configuration, as shown in Fig.~\ref{fig:real_exp}. Initially, the blocks are placed side by side in an "L-T" layout, but their positions and orientations are randomized. This task involves a complex multimodal action distribution, with variability in the sequence of block movements and diverse methods for positioning each block in the desired configuration. Additionally, the task requires the policy to have real-time closed-loop fine control capabilities.}

{We use the Franka Panda robot and define the action space as delta motions along the x and y axes. The policies are trained using 220 human demonstrations, and the experimental results are presented in Table~\ref{tab:real_sr}. The results show that VILP effectively handles this challenging task requiring high-level planning, multimodal action capturing, and precise control. Additionally, VILP enables real-time closed-loop control. In contrast, UniPi performs less effectively on this real-world task and is unable to achieve efficient real-time closed-loop control. Excessive pauses between each action generation result in a highly disjointed action sequence, which adversely affects performance on tasks requiring fine control and dynamic force interaction. Additional videos of these experiments can be found in our supplementary materials.}

\section{Discussion and Future Work}\label{sec:discussion}
We believe that based on VILP, some future directions are very promising.

1. {A promising research direction involves developing a unified robot video generation model capable of generating diverse camera views and tasks. This endeavor may include the development of view-invariant or task-invariant image representations. Such an approach would transcend traditional single-task and single-dataset imitation learning, paving the way for more generalizable and adaptable policies. Additionally, this advancement could significantly enhance the implementation of policies in the wild.}

2. {We demonstrate that video generation robot policies are less dependent on action-labeled data. This finding suggests that future research could explore more generalized methods to translate predicted videos into corresponding actions. In particular, if a universal model for video-to-action mapping is developed, the video generation model (e.g., Sora \cite{brooks2024video}) could assume full responsibility for reasoning and planning tasks. Such a model could be trained on extensive web-scale video data, allowing for the comprehensive utilization of the vast knowledge embedded in such data.}

3. {Considering VILP's feature of video pretraining, it would be intriguing to investigate its ability to capture multi-modal action distributions from cross-domain videos, particularly in longer-horizon tasks that exhibit substantial task-level multi-modality.}

\section{acknowledgements}
This work was partially supported by the National Science Foundation (NSF; Award No. 2423068) and the United States Department of Agriculture (USDA; Awards No. 2023-67021-39072 and 2024-67021-42878). This article solely reflects the opinions and conclusions of its authors and not of NSF or USDA.

\ifCLASSOPTIONcaptionsoff
  \newpage
\fi

\bibliographystyle{IEEEtran}
\bibliography{IEEEabrv,paperref}

\end{document}

%% file: before_document.tex
\usepackage{graphicx}
\usepackage[percent]{overpic}
\usepackage{subfig}
\usepackage{amsmath}
\usepackage{amsthm} 
\usepackage{amssymb}
\usepackage[font=footnotesize]{caption} 
\usepackage{subfig} 
\usepackage[noadjust]{cite} 
\usepackage{color}
\usepackage{algorithm} 
\usepackage{algpseudocode} 
\usepackage{enumerate}
\usepackage[hidelinks,colorlinks=false]{hyperref}
\usepackage[left=0.75in, right=0.75in, top=0.75in, bottom=0.75in]{geometry}
\usepackage{authblk} 
\usepackage{arydshln} 

\usepackage{bm}
\usepackage{tikz}
\usetikzlibrary{calc} 
\usetikzlibrary{shapes} 
\usetikzlibrary{chains}
\usetikzlibrary{fit}
\usetikzlibrary{arrows}
\usetikzlibrary{decorations.text} 
\usetikzlibrary{decorations.markings}
\usetikzlibrary{decorations.pathmorphing} 
\usetikzlibrary{shadows}
\usetikzlibrary{patterns}
\usetikzlibrary{matrix}
\usepackage{pgfplots}
\usepackage[europeanresistors]{circuitikz}
\usepackage[outline]{contour} 
\contourlength{1.5pt}

\graphicspath{{figures/}}

%% file: vilp.bbl
\begin{thebibliography}{10}
\providecommand{\url}[1]{#1}
\csname url@samestyle\endcsname
\providecommand{\newblock}{\relax}
\providecommand{\bibinfo}[2]{#2}
\providecommand{\BIBentrySTDinterwordspacing}{\spaceskip=0pt\relax}
\providecommand{\BIBentryALTinterwordstretchfactor}{4}
\providecommand{\BIBentryALTinterwordspacing}{\spaceskip=\fontdimen2\font plus
\BIBentryALTinterwordstretchfactor\fontdimen3\font minus \fontdimen4\font\relax}
\providecommand{\BIBforeignlanguage}[2]{{%
\expandafter\ifx\csname l@#1\endcsname\relax
\typeout{** WARNING: IEEEtran.bst: No hyphenation pattern has been}%
\typeout{** loaded for the language `#1'. Using the pattern for}%
\typeout{** the default language instead.}%
\else
\language=\csname l@#1\endcsname
\fi
#2}}
\providecommand{\BIBdecl}{\relax}
\BIBdecl

\bibitem{brooks2024video}
T.~Brooks, B.~Peebles, C.~Homes, W.~DePue, Y.~Guo, L.~Jing, D.~Schnurr, J.~Taylor, T.~Luhman, E.~Luhman \emph{et~al.}, ``Video generation models as world simulators,'' 2024.

\bibitem{du2023video}
Y.~Du, M.~Yang, P.~Florence, F.~Xia, A.~Wahid, B.~Ichter, P.~Sermanet, T.~Yu, P.~Abbeel, J.~B. Tenenbaum \emph{et~al.}, ``Video language planning,'' \emph{arXiv preprint arXiv:2310.10625}, 2023.

\bibitem{du2024learning}
Y.~Du, S.~Yang, B.~Dai, H.~Dai, O.~Nachum, J.~Tenenbaum, D.~Schuurmans, and P.~Abbeel, ``Learning universal policies via text-guided video generation,'' \emph{Advances in Neural Information Processing Systems}, vol.~36, 2024.

\bibitem{Ko2023Learning}
P.-C. Ko, J.~Mao, Y.~Du, S.-H. Sun, and J.~B. Tenenbaum, ``{Learning to Act from Actionless Video through Dense Correspondences},'' \emph{arXiv:2310.08576}, 2023.

\bibitem{ge2022long}
S.~Ge, T.~Hayes, H.~Yang, X.~Yin, G.~Pang, D.~Jacobs, J.-B. Huang, and D.~Parikh, ``Long video generation with time-agnostic vqgan and time-sensitive transformer,'' in \emph{European Conference on Computer Vision}.\hskip 1em plus 0.5em minus 0.4em\relax Springer, 2022, pp. 102--118.

\bibitem{yu2023magvit}
L.~Yu, Y.~Cheng, K.~Sohn, J.~Lezama, H.~Zhang, H.~Chang, A.~G. Hauptmann, M.-H. Yang, Y.~Hao, I.~Essa \emph{et~al.}, ``Magvit: Masked generative video transformer,'' in \emph{Proceedings of the IEEE/CVF Conference on Computer Vision and Pattern Recognition}, 2023, pp. 10\,459--10\,469.

\bibitem{yan2021videogpt}
W.~Yan, Y.~Zhang, P.~Abbeel, and A.~Srinivas, ``Videogpt: Video generation using vq-vae and transformers,'' \emph{arXiv preprint arXiv:2104.10157}, 2021.

\bibitem{florence2022implicit}
P.~Florence, C.~Lynch, A.~Zeng, O.~A. Ramirez, A.~Wahid, L.~Downs, A.~Wong, J.~Lee, I.~Mordatch, and J.~Tompson, ``Implicit behavioral cloning,'' in \emph{Conference on Robot Learning}.\hskip 1em plus 0.5em minus 0.4em\relax PMLR, 2022, pp. 158--168.

\bibitem{chi2023diffusionpolicy}
C.~Chi, S.~Feng, Y.~Du, Z.~Xu, E.~Cousineau, B.~Burchfiel, and S.~Song, ``Diffusion policy: Visuomotor policy learning via action diffusion,'' in \emph{Proceedings of Robotics: Science and Systems (RSS)}, 2023.

\bibitem{lee2024behavior}
S.~Lee, Y.~Wang, H.~Etukuru, H.~J. Kim, N.~M.~M. Shafiullah, and L.~Pinto, ``Behavior generation with latent actions,'' \emph{arXiv preprint arXiv:2403.03181}, 2024.

\bibitem{urain2024deep}
J.~Urain, A.~Mandlekar, Y.~Du, M.~Shafiullah, D.~Xu, K.~Fragkiadaki, G.~Chalvatzaki, and J.~Peters, ``Deep generative models in robotics: A survey on learning from multimodal demonstrations,'' \emph{arXiv preprint arXiv:2408.04380}, 2024.

\bibitem{yang2023learning}
M.~Yang, Y.~Du, K.~Ghasemipour, J.~Tompson, D.~Schuurmans, and P.~Abbeel, ``Learning interactive real-world simulators,'' \emph{arXiv preprint arXiv:2310.06114}, 2023.

\bibitem{gu2023seer}
X.~Gu, C.~Wen, W.~Ye, J.~Song, and Y.~Gao, ``Seer: Language instructed video prediction with latent diffusion models,'' \emph{arXiv preprint arXiv:2303.14897}, 2023.

\bibitem{liang2024dreamitate}
J.~Liang, R.~Liu, E.~Ozguroglu, S.~Sudhakar, A.~Dave, P.~Tokmakov, S.~Song, and C.~Vondrick, ``Dreamitate: Real-world visuomotor policy learning via video generation,'' \emph{arXiv preprint arXiv:2406.16862}, 2024.

\bibitem{sohl2015deep}
J.~Sohl-Dickstein, E.~Weiss, N.~Maheswaranathan, and S.~Ganguli, ``Deep unsupervised learning using nonequilibrium thermodynamics,'' in \emph{International conference on machine learning}.\hskip 1em plus 0.5em minus 0.4em\relax PMLR, 2015, pp. 2256--2265.

\bibitem{ho2020denoising}
J.~Ho, A.~Jain, and P.~Abbeel, ``Denoising diffusion probabilistic models,'' \emph{Advances in neural information processing systems}, vol.~33, pp. 6840--6851, 2020.

\bibitem{song2020denoising}
J.~Song, C.~Meng, and S.~Ermon, ``Denoising diffusion implicit models,'' \emph{arXiv preprint arXiv:2010.02502}, 2020.

\bibitem{rombach2022high}
R.~Rombach, A.~Blattmann, D.~Lorenz, P.~Esser, and B.~Ommer, ``High-resolution image synthesis with latent diffusion models,'' in \emph{Proceedings of the IEEE/CVF conference on computer vision and pattern recognition}, 2022, pp. 10\,684--10\,695.

\bibitem{esser2021taming}
P.~Esser, R.~Rombach, and B.~Ommer, ``Taming transformers for high-resolution image synthesis,'' in \emph{Proceedings of the IEEE/CVF conference on computer vision and pattern recognition}, 2021, pp. 12\,873--12\,883.

\bibitem{zhang2018unreasonable}
R.~Zhang, P.~Isola, A.~A. Efros, E.~Shechtman, and O.~Wang, ``The unreasonable effectiveness of deep features as a perceptual metric,'' in \emph{Proceedings of the IEEE conference on computer vision and pattern recognition}, 2018, pp. 586--595.

\bibitem{yu2021vector}
J.~Yu, X.~Li, J.~Y. Koh, H.~Zhang, R.~Pang, J.~Qin, A.~Ku, Y.~Xu, J.~Baldridge, and Y.~Wu, ``Vector-quantized image modeling with improved vqgan,'' \emph{arXiv preprint arXiv:2110.04627}, 2021.

\bibitem{dhariwal2021diffusion}
P.~Dhariwal and A.~Nichol, ``Diffusion models beat gans on image synthesis,'' \emph{Advances in neural information processing systems}, vol.~34, pp. 8780--8794, 2021.

\bibitem{ronneberger2015u}
O.~Ronneberger, P.~Fischer, and T.~Brox, ``U-net: Convolutional networks for biomedical image segmentation,'' in \emph{Medical image computing and computer-assisted intervention--MICCAI 2015: 18th international conference, Munich, Germany, October 5-9, 2015, proceedings, part III 18}.\hskip 1em plus 0.5em minus 0.4em\relax Springer, 2015, pp. 234--241.

\bibitem{jaegle2021perceiver}
A.~Jaegle, F.~Gimeno, A.~Brock, O.~Vinyals, A.~Zisserman, and J.~Carreira, ``Perceiver: General perception with iterative attention,'' in \emph{International conference on machine learning}.\hskip 1em plus 0.5em minus 0.4em\relax PMLR, 2021, pp. 4651--4664.

\bibitem{vaswani2023attention}
A.~Vaswani, N.~Shazeer, N.~Parmar, J.~Uszkoreit, L.~Jones, A.~N. Gomez, L.~Kaiser, and I.~Polosukhin, ``Attention is all you need,'' 2023.

\bibitem{xu2024unit}
\BIBentryALTinterwordspacing
Z.~Xu, R.~Uppuluri, X.~Zhang, C.~Fitch, P.~G. Crandall, W.~Shou, D.~Wang, and Y.~She, ``{UniT}: Unified tactile representation for robot learning,'' 2024. [Online]. Available: \url{https://arxiv.org/abs/2408.06481}
\BIBentrySTDinterwordspacing

\bibitem{xu2024leto}
Z.~Xu and Y.~She, ``Leto: Learning constrained visuomotor policy with differentiable trajectory optimization,'' \emph{arXiv preprint arXiv:2401.17500}, 2024.

\bibitem{zeng2021transporter}
A.~Zeng, P.~Florence, J.~Tompson, S.~Welker, J.~Chien, M.~Attarian, T.~Armstrong, I.~Krasin, D.~Duong, V.~Sindhwani \emph{et~al.}, ``Transporter networks: Rearranging the visual world for robotic manipulation,'' in \emph{Conference on Robot Learning}.\hskip 1em plus 0.5em minus 0.4em\relax PMLR, 2021, pp. 726--747.

\bibitem{mandlekar2022matters}
A.~Mandlekar, D.~Xu, J.~Wong, S.~Nasiriany, C.~Wang, R.~Kulkarni, L.~Fei-Fei, S.~Savarese, Y.~Zhu, and R.~Mart{\'\i}n-Mart{\'\i}n, ``What matters in learning from offline human demonstrations for robot manipulation,'' in \emph{Proc. Conf. Robot Learn.}, 2022, pp. 1678--1690.

\bibitem{chi2022iterative}
C.~Chi, B.~Burchfiel, E.~Cousineau, S.~Feng, and S.~Song, ``{Iterative Residual Policy}: for goal-conditioned dynamic manipulation of deformable objects,'' in \emph{in Proc. of Robot.: Sci. and Syst}, 2022.

\end{thebibliography}
